\documentclass{article}
\usepackage{graphicx} % Required for 
\usepackage{subcaption}
\usepackage[numbers, sort&compress]{natbib} 
\usepackage{array} % Optional, for more table features

\title{BioAgents: Democratizing Bioinformatics Analysis with Multi-Agent Systems}

\author{Nikita Mehandru, Amanda K. Hall, Olesya Melnichenko, \and   Yulia Dubinina, Daniel Tsirulnikov, David Bamman,  
\and Ahmed Alaa, Scott Saponas,  Venkat S. Malladi}

\date{}  %remove the date

\begin{document}

\maketitle

\begin{abstract}
Creating end-to-end bioinformatics workflows requires diverse domain expertise, which poses challenges for both junior and senior researchers as it demands a deep understanding of both genomics concepts and computational techniques. While large language models (LLMs) provide some assistance, they often fall short in providing the nuanced guidance needed to execute complex bioinformatics tasks, and require expensive computing resources to achieve high performance. We thus propose a multi-agent system built on small language models, fine-tuned on bioinformatics data, and enhanced with retrieval augmented generation (RAG). Our system, BioAgents, enables local operation and personalization using proprietary data. We observe performance comparable to human experts on conceptual genomics tasks, and suggest next steps to enhance code generation capabilities.

\end{abstract}

\section{Main}

Large language models (LLMs) have been applied to various domain-specific contexts, including scientific discovery in medicine \cite{thirunavukarasu2023large, singhal2023large, yang2023large}, chemistry \cite{bran2023chemcrow, boiko2023autonomous}, and biotechnology \cite{madani2023large}. 
Recent advances in LLMs have spurred their use in bioinformatics~\cite{cheng2024l2g}, a field encompassing data-intensive tasks such as genome sequencing, protein structure prediction, and pathway analysis. One of the most significant applications has been AlphaFold3, which uses transformer architecture with triangular attention to predict a protein's three-dimensional (3-D) structure from amino acid sequences~\cite{abramson2024accurate}. Other applications include the use of protein language models in transforming amino acids into embeddings \cite{yin2024evaluation}.

While LLMs demonstrate impressive capabilities, these models have been found to struggle on complex genomics~\cite{bhardwaj2023chatgpt, lubiana2023ten, sarwal2023biollmbench} and bioinformatics code generation~\cite{wang2024large, kang2024integrating, tang2024biocoder} tasks with their performance and time to arrive at the correct solution varying significantly with task complexity. For example, ChatGPT solved 97.3\% of programming exercises from an introductory bioinformatics course within seven or fewer tries~\cite{piccolo2023many}. The model; however, was only able to solve 75.5\%, or 139 out of 184 exercises, on its first attempt. This disparity highlights that while LLMs can assist bioinformatics researchers with introductory data analysis questions, they encounter challenges when tackling more intricate and complex real-world programming, analysis, and research questions, often requiring knowledge of multiple tools, data formats, and analysis techniques.

\begin{figure}[tbh]
  \centering
  \includegraphics[width=\textwidth]{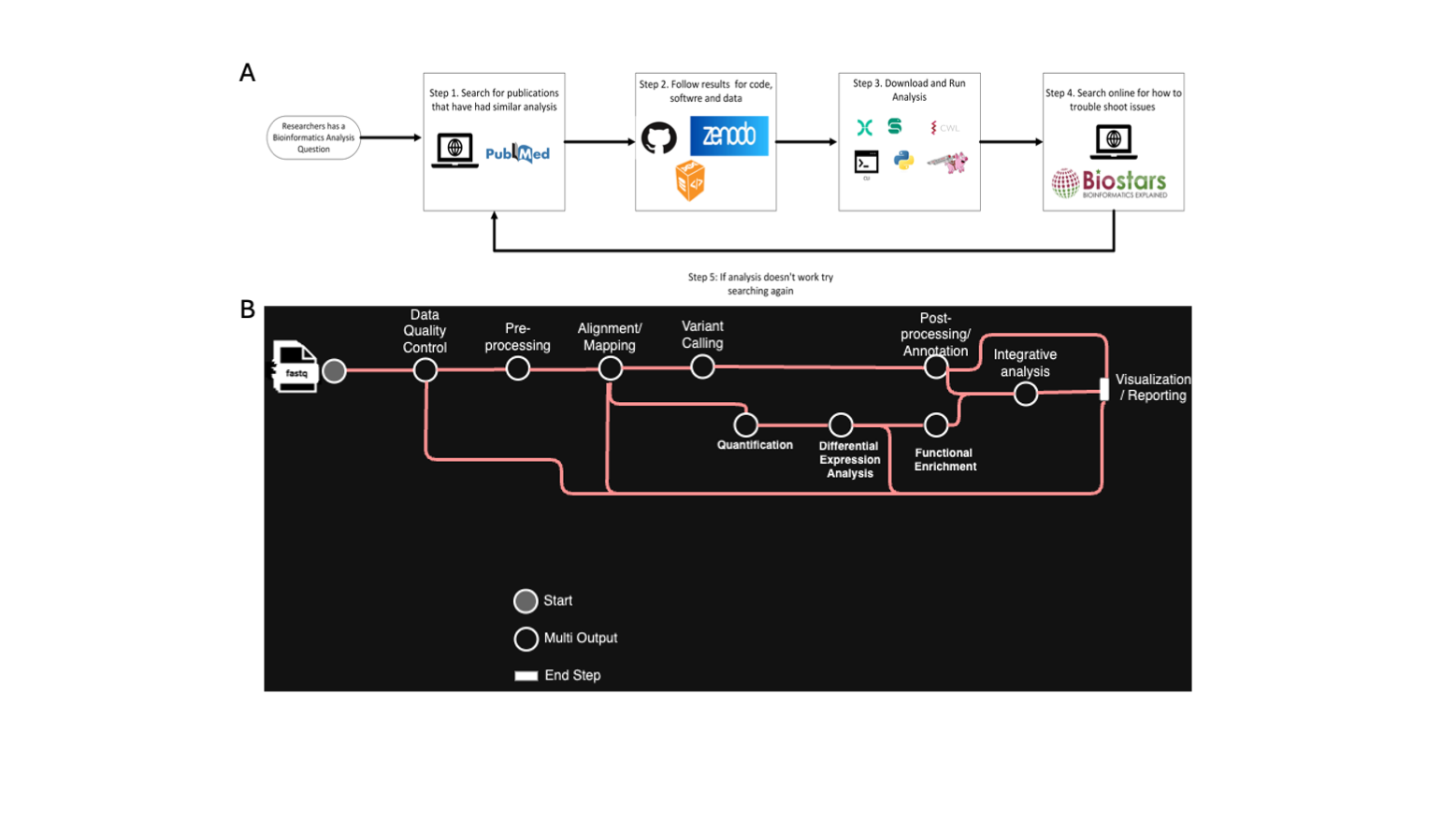}
  \caption{\textbf{Supporting Research through Knowledge Graphs and Directed Acyclic Graphs.} A. A knowledge graph showcases the current state-of-the-art for a researcher to start from a research question and navigate through relevant data, tools, and methods to independently run their analysis. B. A typical bioinformatics workflow represented as a Directed Acyclic Graph (DAG), showcasing the intricate dependencies between tasks such as data preprocessing, genome assembly, annotation, and analysis, where each node represents a computational step and edges indicate the flow of data or control.}
  \label{generic_workflow}
\end{figure}

\sloppy
A frequent challenge for bioinformatics researchers is navigating the complexity of building end-to-end pipelines, which typically requires expertise across multiple domains. Bioinformaticians often mine question-answer platforms like Biostars for similar problems, search for reproducible scientific workflow examples (e.g., Nextflow, WDL or Snakemake) and software containers (e.g., Biocontainers) on GitHub~\cite{parnell2011biostar, di2017nextflow, gruening2019recommendations}, or refer to the methods sections of recently published papers for code. The creation of these workflows require several key steps involving various dependencies, software, compute, storage, data, and vast expertise, including data pre-processing, alignment, and post-processing (as shown in Figure~\ref{generic_workflow}). This complexity can present a steep learning curve for newcomers, and poses challenges for bioinformatics experts to stay up-to-date with new techniques~\cite{shue2023empowering, chatterjee2018guide}, as well as with analysis-specific software versions. While established open-source community platforms provide one-off exchanges, they offer limited guidance for researchers trying to develop complex, multi-step workflows across a network of on-premise and cloud infrastructure~\cite{patel2023beginner}. As a result, there is a need for interactive and dynamic tools that can offer continuous support. 

To bridge this gap and democratize access to bioinformatics knowledge, we introduce BioAgents~-- a multi-agent system designed to assist users in designing, developing, and troubleshooting complex bioinformatics pipelines. Recognizing the potential of multi-agent frameworks~\cite{wu2023autogen, li2023camel, singh2024revolutionizing, sreedhar2024simulating}, our system provides an interactive solution that adapts to the ongoing needs of users working in specialized domains~\cite{mehandru2024evaluating, wang2024survey, xiao2024cellagent, gao2024empowering, xin2024bioinformatics}.

\begin{figure}[tbh]
    \centering
    \vspace{-3mm}
    % Create a larger figure with all three panels in a single row
    \begin{subfigure}[b]{0.45\textwidth}
        \centering
        \includegraphics[width=\textwidth]
        {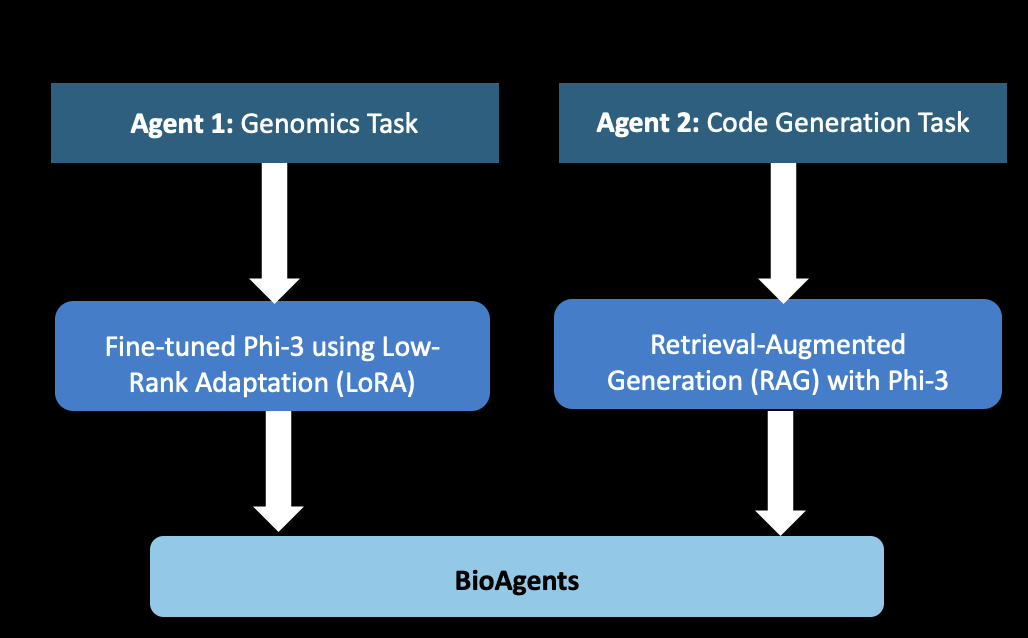}
        \caption{}
        \label{exp_design}
    \end{subfigure}
    \hfill % Horizontal space between subfigures
    \begin{subfigure}[b]{0.31\textwidth}
        \centering
        \includegraphics[width=\textwidth]{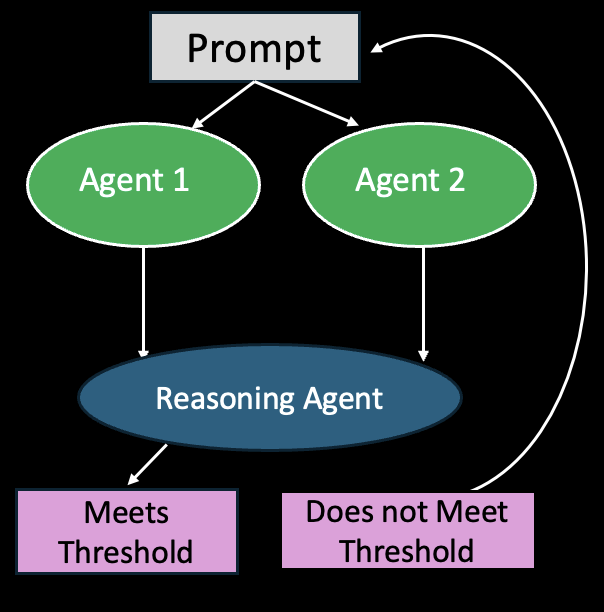}
        \caption{}
        \label{final_agent}
    \end{subfigure}
    \hfill % Horizontal space between subfigures
    \begin{subfigure}[b]{0.55\textwidth}
        \centering
\includegraphics[width=\textwidth]{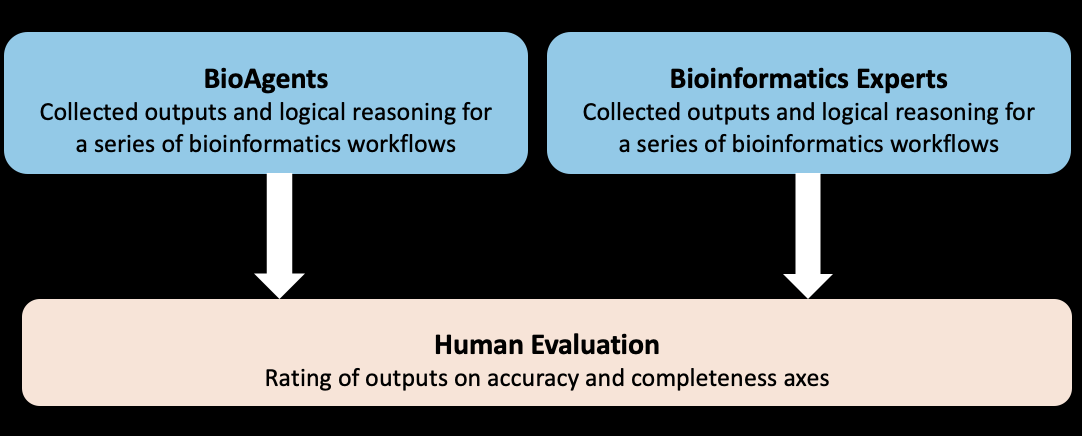}
        \caption{}
        \label{comparison}
    \end{subfigure}
    
   \caption{(a) \textbf{Two Specialized Agents.} Each specialized agent used Phi-3. The first agent focused on conceptual genomics tasks and was fine-tuned on bioinformatics tools documentation, while the second agent used retrieval-augmented generation (RAG) on workflow documentation. (b) \textbf{Overview of BioAgents.} The reasoning agent, a baseline Phi-3 model, processes the outputs from each specialized agent independently and generates the final response. (c) \textbf{Comparison of BioAgents' Outputs with Expert Outputs.}}
    \label{fig:all_panels}
\end{figure}

To better understand the challenges faced by practitioners, we analyzed 68,000 question-answer (QA) pairs from Biostars, extracting the associated tags and categorizing each question. The most frequent questions on the platform revolved around tools, specifically bioinformatics software programs and packages, as well as analysis, such as pipeline-related queries focused on RNA-sequencing, alignment, and variant calling. To address the diverse and complex nature of these questions, we employ multiple specialized agents, each tailored to handle specific tasks such as tool selection, workflow generation, and error troubleshooting, enabling a modular and efficient approach to solving bioinformatics challenges. These insights directly informed the development of our two specialized agents within BioAgents. 

While existing multi-agent systems primarily rely on large language models~\cite{m2024augmenting, guo2023can}, we leveraged a smaller, more efficient language model, Phi-3~\cite {abdin2024phi}. By using a smaller language model, we are able to maintain high performance while significantly reducing computational resources and infrastructure~\cite{schick2024toolformer, chen2024role}. Avoiding the heavy infrastructure demands associated with larger models, BioAgents is more accessible for local use and efficient real-time applications. 

%LLMs are computationally demanding. Due to the high dimensionality and complexity of biological and genomic data, techniques such as fine-tuning and retrieval-augmented generation (RAG) can improve model performance, but require substantial computational resources. Powerful graphics processing units (GPUs), large memory capacities, and intensive pre-processing required for fine-tuning hyperparameters all contribute to the computational burden. As a result, these techniques are often expensive to execute and require considerable infrastructure. In contrast, small language models offer a promising solution allowing researchers to run models locally without requiring significant computational resources~\cite{schick2024toolformer, chen2024role} and infrastructure.

We used the baseline Phi-3 model to build three agents: two specialized agents and Phi-3 as the reasoning agent. Our first agent focused on conceptual genomics tasks, and was fine-tuned on bioinformatics tools documentation from Biocontainers and the software ontology ~\cite{da2017biocontainers, malone2014software}. Our second agent used retrieval-augmented generation (RAG) on nf-core documentation and the EDAM ontology \cite{ewels2020nf, jon_ison_2020_3899895, black2022edam}. Figure~\ref{fig:all_panels} shows the creation of our two specialized agents, an overview of the BioAgents, and our experimental design, respectively. 

%code pre-training datasets don't train on docker files:
%%%%https://huggingface.co/datasets/bigcode/the-stack-v2
%%%% importance of error recovery in LLM agents. 
%%%% self-correcting
%sandboxing- limiting execution environment.
%credentialing- limiting llm access.
%post-hoc auditinh- generate actions and can decide whether to execute. 
%%%% see lecture for next steps////future directions: human in the loop, agentic training methods. 

\section{Results}

%Results and Methods should be divided by topical subheadings; the Discussion does not contain subheadings.
\subsection{Evaluation across several use  cases}

We devised three use cases of varying difficulty to evaluate our multi-agent system. These workflows, listed below, were designed to assess both conceptual genomics (analysis steps) and code generation tasks. We recruited bioinformatics experts, and provided them with the same inputs used by the multi-agent system. Each workflow involved completing the conceptual genomics and code generation tasks, providing any additional information needed to aid in answering the user query, and explaining the logical reasoning behind the final output.

\vspace{0.4cm}
\textbf{Conceptual Genomics and Code Generation Tasks} 

\vspace{0.1cm}
\textit{Level 1 Tasks (Easy)}
\begin{itemize}
    \item How would I provide quality metrics on FASTQ files?
    \item What code or workflow do I need to write to provide quality metrics on FASTQ files?
\end{itemize}

\textit{Level 2 Tasks (Medium)}
\begin{itemize}
    \item How do I align RNA-seq data against a human reference genome?
    \item What code or workflow do I need to write to align RNA-seq data against a human reference genome?
\end{itemize}

\textit{Level 3 Tasks (Hard)}
\begin{itemize}
    \item How can I assemble, annotate, and analyze SARS-CoV-2 genomes from sequencing data to identify and characterize different variants of the virus?
    \item What code or workflow do I need to write to assemble, annotate, and analyze SARS-CoV-2 genomes from sequencing data to identify and characterize different variants of the virus?
\end{itemize}

\begin{figure}[tbh]
  \centering
\fbox{\includegraphics[width=0.85\textwidth]{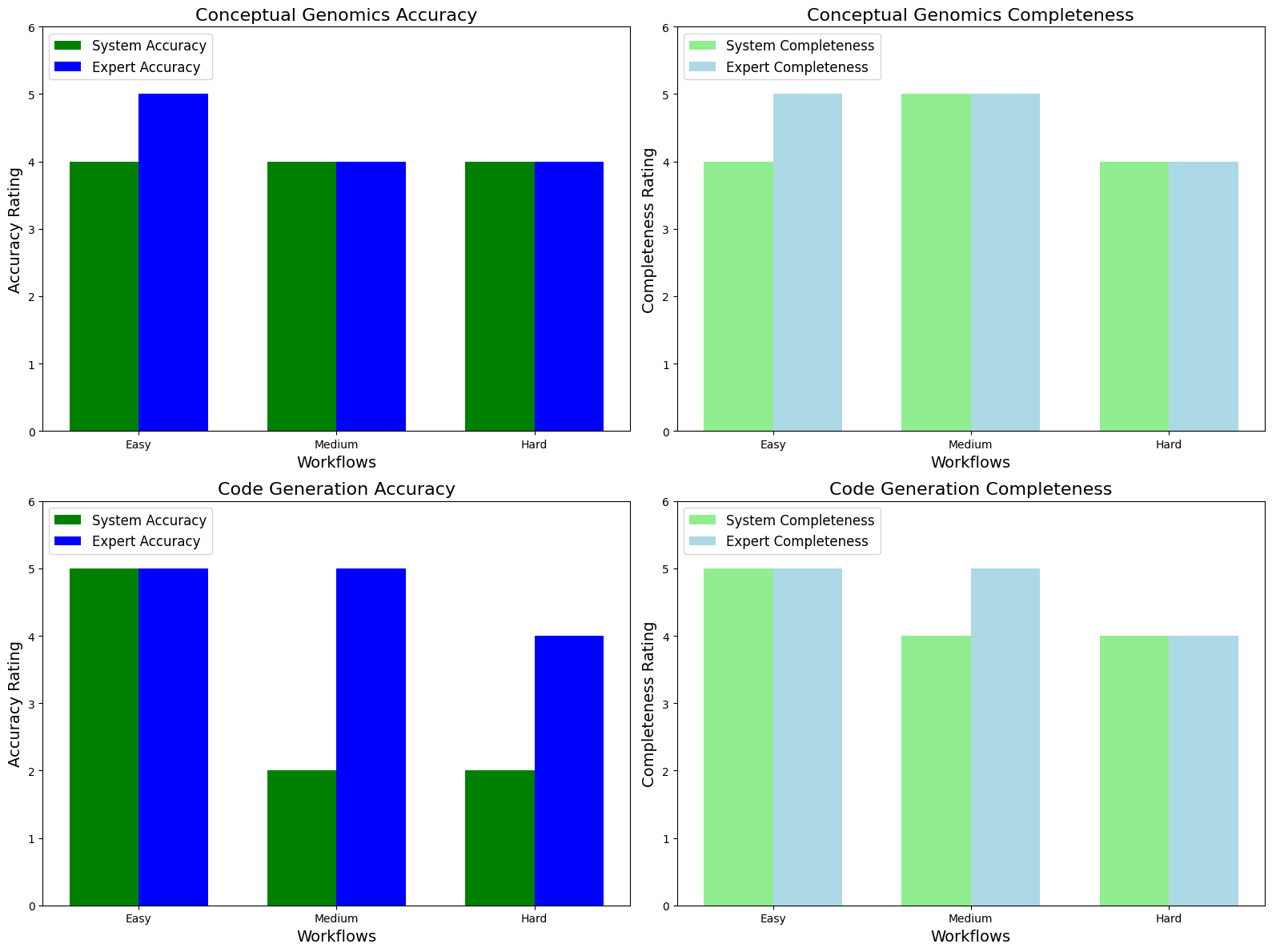}}
  \caption{\textbf{Comparison of system and expert performance across conceptual genomics and code generation tasks.}The top row evaluates conceptual genomics tasks, with separate panels for accuracy (left) and completeness (right). The bottom row evaluates code generation tasks, similarly split into accuracy (left) and completeness (right). For conceptual genomics tasks, the system demonstrates comparable performance to human experts across all levels of difficulty. In code generation tasks, the system matches expert performance on easier tasks, but shows a decline in accuracy and completeness for medium and hard tasks, highlighting opportunities for improvement in addressing complex challenges.}
  \label{main_results}
\end{figure}

To assess performance, a bioinformatician reviewed both the system and human expert outputs on two axes: 1) accuracy, and 2) completeness. Accuracy was defined as how well the user's query was answered, while completeness referred to the extent to which the output captured all relevant information in response to the user query. Evaluation results are presented in Figure~\ref{main_results}.

\subsubsection{Conceptual Genomics Tasks}

BioAgents demonstrated performance on par with experts on conceptual genomics questions across three workflows. This success is largely attributed to our use of Low-Rank Adaptation (LoRA) to fine-tune an agent on the top 50 bioinformatics tools in Biocontainers, including detailed software versions and help documentation. Biocontainers, a widely used bioinformatics service, provides the infrastructure for managing bioinformatics packages and containers, such as conda and docker. Consequently, the system effectively interpreted and responded to these conceptual tasks, achieving human-expert-like performance.

\begin{figure}[ht]
  \centering
  \begin{minipage}{0.48\textwidth}
    \centering
    \fbox{\includegraphics[width=\textwidth]{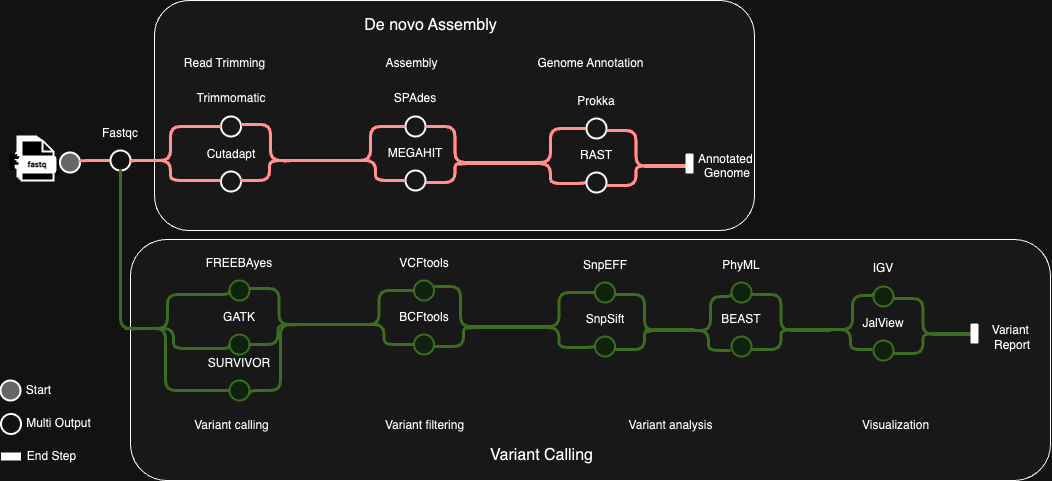}}
    \caption*{(a) BioAgents on the Hard Workflow, broken down into two key parallel workflows, De novo Assembly and Variant calling.} % Caption without label
  \end{minipage}
  \hfill
  \begin{minipage}{0.48\textwidth}
    \centering
    \fbox{\includegraphics[width=\textwidth]{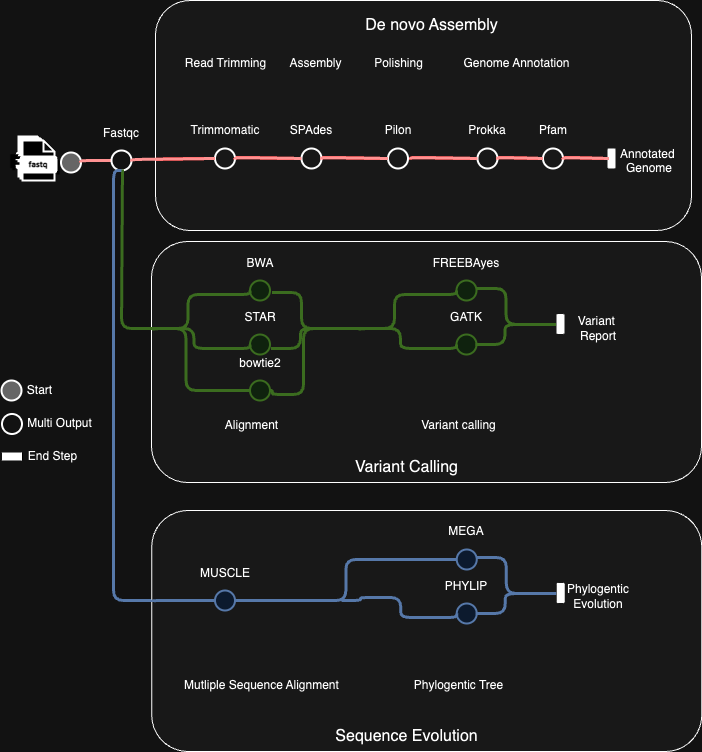}}
    \caption*{(b) Experts on the Hard Workflow, broken down into three parallel workflows, De novo Assembly, Variant calling, and Sequence evolution.} % Caption without label
  \end{minipage}
  \caption{BioAgents and Experts on the Hard Genomics Workflow}
  \label{fig:hard_workflow}
\end{figure}

In a challenging workflow question on assembling, annotating, and analyzing SARS-CoV-2 genomes from sequencing data, BioAgents provided a logical series of steps, including obtaining sequencing data, performing quality control, assembling the high-quality reads using a de novo assembler, annotating the assembled genome using tools like Prokka or RAST, identifying and characterizing variants, and constructing a phylogenetic tree \cite{seemann2014prokka, aziz2008rast}. While human experts proposed robust pipelines, they lacked rationales for tool recommendations. BioAgents did occasionally omit steps though, requiring users to fill in the gaps as shown in Figure~\ref{fig:hard_workflow}.

\subsubsection{Code Generation Tasks}

Performance discrepancies emerged in code generation tasks, particularly in workflows of increasing complexity. For easy tasks, BioAgents matched expert accuracy, but sometimes provided false information about tools. For medium tasks, representing end-to-end pipelines like those in nf-core workflows (https://nf-co.re/pipelines/), BioAgents struggled to produce complete outputs. In the most complex workflows, the system failed to generate starter code, instead offering step outlines more similar to a conceptual answer. These limitations were attributed to gaps in the indexed workflows, and a lack of tool and language diversity in the training dataset.

\subsection{Reliability and Transparency}

Two key components are necessary in the deployment of multi-agent systems in highly specialized domains: reliability and transparency. Reliability ensures that the system consistently delivers accurate results, while transparency enables users to understand and trust the system's decision-making process. 

\subsubsection{Self-Reflection in Agent Systems}

Several techniques have been proposed to enable a language model to correct its outputs based on internal evaluation, including: self-consistency \cite{chen2023two}, self-correction~\cite{pan2023automatically, kamoi2024can}, self-evolution~\cite{tao2024survey}, self-feedback~\cite{liang2024internal} and self-evaluation~\cite{ren2023self}. 

BioAgents incorporated self-evaluation to enhance output reliability, inspired by the idea that agent systems can assess the accuracy of their own outputs~\cite{zhuge2024agent}. Our reasoning agent assessed the quality of responses against a defined threshold. Outputs scoring below this threshold were reprocessed, with agents independently reanalyzing the prompts before returning results. However, the iterative process revealed diminishing returns, where repeated refinements negatively impacted output quality and might not necessarily lead to improved outcomes.

\begin{figure}[ht]
  \centering
\fbox{\includegraphics[width=0.85\textwidth]{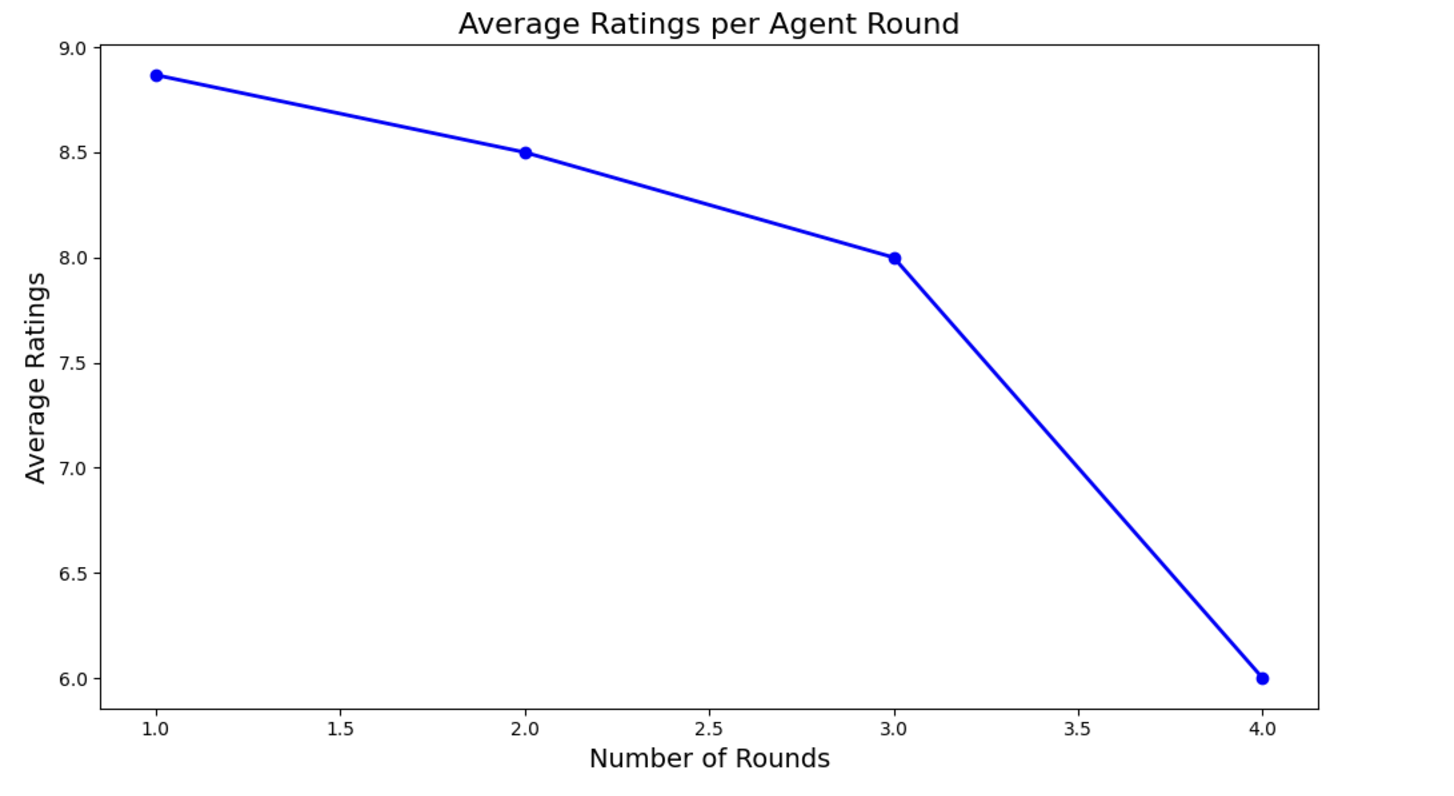}}
  \caption{Self-Ratings by Number of Rounds: an inverse correlation between the number of rounds the multi-agent system takes to reach the final answer and the quality of the output's rating suggests a potential limitation of the iterative processes in multi-agent systems.}
  \label{agents_per_round}
\end{figure}

\subsubsection{Collaborative Reasoning and Transparent Guidance}

Current applications of LLM agents in domain-specific tasks have struggled in the areas of long-term reasoning, decision-making, and instruction-following~\cite{liu2023agentbench}. Moreover, their proficiency in addressing practical bioinformatics queries and conducting nuanced knowledge inference remains constrained~\cite{chen2023bioinfo}.

In our experimental set-up, both the system and human experts were asked to explain any additional information they would need to better answer users' questions, and the logical reasoning process behind their answers. Our motivation behind this assessment stemmed from various existing reasoning frameworks~-- chain-of-thought (CoT)~\cite{wei2022chain} and ReAct~\cite{yao2022react}~-- used to provide interpretability in LLMs. 

In the context of the conceptual genomics medium workflow task, BioAgents explained its rationale for selecting the most suitable RNA-seq alignment tool for mapping against the reference genome, recommending STAR and HISAT2 for their high-throughput and accurate alignments \cite{dobin2013star, kim2019graph}. The multi-agent system described how these alignment tools mapped the RNA-seq reads to the reference genome, enabling the identification of genomic locations within the read. BioAgents also specified the factors influencing its choice of alignment tool, noting the size of the RNA-seq dataset and the user's desired accuracy level as important factors. Generating natural language explanations of model outputs has been shown to improve interpretability, thereby increasing transparency and fostering trust, as users are better able to understand the reasoning behind the model's specific outputs~\cite{brown2024enhancing, schwartz2023enhancing}.

A key insight from our findings was that our multi-agent system was able to identify additional information that would have improved responses, even in cases where accuracy was lower. For example, in the hard code generation task, BioAgents struggled to generate the necessary workflow code, but was able to identify additional information that could have better answered the user question, specifically more information on the raw sequencing data, the reference genome sequence, software and tool versions, computational resources (e.g., CPU, memory, disk space), and the user's bioinformatics experience. In contrast, although the human experts achieved a higher accuracy score on the same task (four compared to two), they described the limitations of their responses, with one human expert noting that the solutions provided were "cobbled together from searching for tutorials," indicating that it was difficult to identify their information gaps. BioAgents, on the other hand, demonstrated more metacognitive awareness recognizing what it didn't know, and more importantly, additional information it could benefit from~\cite{didolkar2024metacognitive}. A consistent theme across the human expert responses was an inability to articulate what additional information they would need to improve their answers. This highlights how vast and expansive of a knowledge base is needed to fully answer these questions, which often extends beyond the expertise of one individual.
%limitation 
\section{Discussion}

The reproducibility crisis in computational research highlights the urgent need for systems that can reliably extract, reproduce, and adapt research findings. This challenge is especially pronounced in bioinformatics, where complex workflows often hinder replication and validation efforts. BioAgents, a multi-agent system, offers a promising solution by extracting methods from research papers, generating executable workflows, and integrating human-in-the-loop approaches to improve accuracy and customization.

BioAgents has the potential to facilitate reproducibility by automatically synthesizing workflows from research publications, enabling researchers to replicate experiments, validate results, and adapt analyses to their datasets. By prioritizing transparency and integrating human feedback, BioAgents ensures outputs are both reliable and user-specific. Furthermore, BioAgents can be extended to clinical settings and other scientific domains. In medicine, the system could assist in replicating diagnostic workflows, personalizing treatment recommendations based on patient data, and optimizing clinical trial designs, ultimately enabling more efficient and reliable translational research. In chemistry and physics, BioAgents can automate the replication of experimental protocols and model complex systems, enhancing the reproducibility of results across a wide range of scientific fields.

Collaborative reasoning highlighted key areas for improvement in reasoning, decision-making, and instruction-following. BioAgents effectively identified information gaps, such as tool versions and user experience, which experts often overlooked. Generating natural language explanations of its decisions increased interpretability, fostering user trust. Despite struggles with accuracy in complex tasks, BioAgents demonstrated metacognitive awareness, outlining additional data that could improve results.

The increased reliability, transparency, and trust fostered by BioAgents is particularly valuable for reducing barriers for new bioinformatics researchers. Unlike static question-answer forums, which can provide solutions without clear explanations or insight into the respondent's reasoning process, BioAgents allows users to not only receive answers, but also provides the underlying supporting information that led to those answers. By sharing the multi-agent system's reasoning behind its proposed steps, researchers can learn how to replicate those decision-making processes, highlighting the educational value of our approach. Ultimately, the transparency provided by BioAgents not only improves trust in agentic systems but also facilitates knowledge transfer, allowing users to grow and develop their expertise. 

In conceptual genomics tasks, BioAgents demonstrated performance comparable to that of human experts, successfully addressing domain-specific challenges. However, areas for improvement were identified in code generation. Specifically:

\begin{itemize}
    \item \textbf{Workflow Scope}: The system's reliance on nf-core workflows limited diversity. Expanding indexed workflows to include additional sources could address this gap.
    \item \textbf{Information Retrieval}: Retrieving multiple document matches, rather than relying solely on top-ranked results, could enhance the relevance of generated workflows.
    \item \textbf{Reasoning Agent}: Enhancing the reasoning agent to verify tool versions, ensure executability, and reference source documentation could increase transparency and foster user trust.
\end{itemize}

One of BioAgents' key strengths is its ability to support user learning. By linking generated workflows to source documentation and providing support information for each step, BioAgents enables researchers to understand and modify workflows, thereby contributing to their professional development and the broader bioinformatics community. Additionally, when BioAgents requests supplementary information, researchers gain the opportunity to refine results tailored to their specific analyses, while also enhancing their understanding of their own data.

By lowering barriers to compute resources and operating seamlessly in local environments, BioAgents addresses both accessibility and scalability. Its potential extends beyond bioinformatics, offering a model for intelligent systems in other domains facing reproducibility challenges to use the BioAgents framework and train on their own propriety or domain-specific code and documentation. 

With targeted enhancements in workflow diversity, retrieval methods, and reasoning capabilities, BioAgents is poised to become a cornerstone in the push for reproducible, transparent, and accessible computational research.
\section{Methods}

\subsection{Datasets}

\subsubsection{Biostars}
Biostars~\cite{parnell2011biostar} is an online community platform for the bioinformatics community where users can ask and answer questions related to computational genomics and biological data analysis. We scraped all publicly available data from the site, which included a total of 68,000 question-answer (QA) pairs, up to May~1,~2024. Only answers with atleast one user upvote were added to the QA dataset. Tags assigned to each question were extracted, then GPT-3.5 was used to categorize them into one of five categories: 
\begin{itemize}
    \item tool~-- software programs and packages used for bioinformatics analysis;
    \item analysis~-- pipelines and analysis performed in bioinformatics field, such as rna-seq, alignment, variant calling;
    \item data format~-- genomics and other -omics data formats;
    \item programming~-- programming laguages, including wdl, nextflow and snakemake, and operation systems;
    \item other~-- for everything else.
\end{itemize}

\subsubsection{Biocontainers}

We fine-tuned BioAgents on Biocontainers’ (https://biocontainers.pro/) top 50 tools, including versions and documentation. We use the TRS API (https://api.biocontainers.pro/ga4gh/trs/v2/ui/) to pull statistics on the top bioinformatics tools, based on download frequency. Then, we grabbed each available docker version of each of those tools. For each docker container, we downloaded the container and outputted command-line help documentation. 

\subsubsection{Ontologies}

We downloaded both the Software  and EDAM Ontologies ~\cite{malone2014software,jon_ison_2020_3899895, black2022edam} for software and assay description. To convert to JSON-LD we used the JSON or OBO format to extract the name and either the description or definition .

\subsection{Models}
We leveraged a single A100 GPU to perform parameter-efficient fine-tuning (PEFT) on the Phi-3-mini-128-instruct model, optimizing it for bioinformatics tasks. Specifically, we employ the QLoRA technique, which enables fine-tuning with reduced computational overhead by quantizing the model's layers and training low-rank adapters. This approach is particularly well-suited for large-scale language models like Phi-3-mini, as it retains performance while significantly reducing resource requirements.

Our fine-tuning dataset focuses on the top 50 most commonly used BioContainers tools, along with their associated versions and help documentation, ensuring broad applicability to bioinformatics workflows. Additionally, we added Software Ontology data about the name of the software and its purpose. Training was conducted on Azure Machine Learning, with model configurations limited to a new token count of 1,000 and a temperature of 0.1 to control response diversity and precision.

For our retrieval-augmented generation (RAG) implementation, we integrate OpenAI's text-embedding-ada-002 for high-quality semantic search. The embeddings are indexed within Azure AI's search service, optimized to retrieve nf-core modules efficiently, and the Sequence Ontology, describing each assay and its purpose. This combination ensures that the system can provide relevant, tool-specific code generation and guidance tailored to bioinformatics workflows.

By combining QLoRA-based fine-tuning and RAG, we achieve a system that balances computational efficiency, domain specificity, and accessibility for researchers in bioinformatics. 

\subsection{Expert Survey}

We conducted a survey to elicit expert bioinformatician responses to workflow questions informed by Biostars data, and evaluated this against outputs from BioAgents. Our survey assessed human expert’s reasoning and logic behind their responses. Below we discuss our survey design, respondent recruitment, informed consent process, survey data analysis, and findings.  

\subsubsection{Survey Design}

We created a survey using Microsoft Forms to obtain human expert answers and their reasoning behind responses related to translating genomics tasks, and writing subsequent code to analyze data question types. Biostars community forum questions were abstracted and categorized to evaluate common question types, which we found to be around tools and/or analysis. We then created three levels of questions (easy, medium, and hard) increasing in complexity (i.e., number of steps and knowledge required) derived from questions in Biostars. 

The survey consisted of 27 questions. We asked eight demographic questions to assess respondent’s education-level, number of years working with bioinformatics tools, data types they worked with regularly, work setting, programming experience, age, gender, and if English was their first language. For the remaining survey questions, respondents were asked to imagine an undergraduate student asked for help on a set of questions, and to provide the logic and steps they would advise the student to take to successfully complete their inquiry. Each of the three question levels had two parts: 1) answering the question asked by the student with corresponding logic/ reasoning, and 2) generating code (in their preferred programming language) with corresponding logic /reasoning. An optional question was asked after each of these questions around what additional information, if any, they would need to answer the student’s question.  

\subsubsection{Recruitment and Informed Consent}

We recruited survey respondents through a combination of Microsoft internal community distribution email lists and snowball sampling techniques in August of 2024. The primary inclusion criteria for participation required individuals to have 5+ years of bioinformatics experience with tools/ workflows, intermediate to advanced coding skills, and be at least 18 years of age. 

Five bioinformatician experts responded to our survey. The expert respondents ranged in age (25-44 years), gender (2 women, 3 men), and education level (2 masters, 3 doctorates). All respondents reported intermediate to advanced programming experience, 5 or more years of experience working with bioinformatics tools, and English as their first language. Additionally, all respondents reported working with a range of biomedical data types (3 non-human, 3 human, non-clinical, and 4 human, clinical data) and most reported working in an industry setting.   

Respondents were excluded if they had higher executive roles in industry to avoid any potential bias. Respondents interested in the study were sent the Microsoft Forms survey to assess their eligibility based on the study criteria, and to review the study informed consent. If they met the eligibility criteria requirements and agreed to the terms outlined in the informed consent, they proceeded to respond to the survey questions. Respondents were compensated \$50 USD for their time via a gift card. This study was reviewed and approval by Microsoft Research Institutional Review Board (ID10950). 

\subsubsection{Qualitative Data Analysis}

We conducted a content and thematic analysis of the open-ended questions. For the three question levels (easy, medium, hard) related to the genomics workflows and code generation tasks, we conducted a content analysis of the steps and code across respondents to create a ground-truth dataset to evaluate against the output of BioAgents. For the logic/reasoning and additional information needed questions we conducted a thematic analysis of emerging themes.  

\subsubsection{Expert Findings}

We aggregated survey responses from the five expert respondents on Agent 1 and Agent 2 questions across the three levels of difficulty (Figure 2). We created a master list of steps by task, and corresponding code as ground-truth data across all experts’ responses to the genomics workflow and code generation questions. We then conducted a thematic analysis for emerging themes based on responses to the logic/reasoning and additional information needed questions. 

Expert responses to questions and corresponding logic plus additional information required did not vary across experts on the easy question types. However, for the medium and hard question sets, experts required more logic/reasoning and additional information due to the increased complexity (e.g., number of steps) of the biomedical research tasks necessary to correctly answer the question. This also included the tools and knowledge needed to navigate and implement the correct solution.

\subsection{Human Evaluation}

 To assess the performance of the system and experts, we conducted a human evaluation study in which a domain expert scored outputs on two key criteria: \textbf{accuracy} and \textbf{completeness}.  For conceptual tasks, experts assessed whether the system's reasoning and recommendations were consistent with domain knowledge. For code generation tasks, experts verified syntax correctness, tool compatibility, and functionality. 
\begin{enumerate}
    \item \textbf{Accuracy (1-5):} The degree to which the output code and steps are correct. The steps and/or software are reasonable and not hallucinated. Where 1 indicates major inaccuracies and 5 indicates full correctness.
    \item \textbf{Completeness (1-5):} The extent to which the output provides all necessary components or steps required for the task, where 1 indicates significant omissions and 5 indicates comprehensive coverage of code or steps such that the user would be able to implement the answer without searching for additional information.
\end{enumerate}

\subsection{Supplementary Material}
%used semantic ranker: uses deep neural networks to provide relevant results based on semantics not just lexical analysis. see more here: https://learn.microsoft.com/en-us/azure/search/semantic-search-overview

Our benchmarking results indicate that Phi-3 and GPT-4 perform similarly on Biostars QA pairs. 

\begin{table}[h]
    \centering
    \begin{tabular}{|c|p{2cm}|p{2cm}|p{2cm}|p{2cm}|}
        \hline
        \textbf{Model} & \textbf{ROUGE-1} & \textbf{ROUGE-2} & \textbf{ROUGE-L} & \textbf{ROUGE-L-SUM} \\ \hline
        Phi-3.5-mini & 0.129 & 0.015 & 0.074 & 0.092 \\ \hline
        Phi-3.5-MoE & 0.129 & 0.015 & 0.074 & 0.091 \\ \hline
        GPT-4 & 0.183 & 0.029 & 0.103 & 0.125 \\ \hline
        GPT-4o & 0.122 & 0.014 & 0.072 & 0.091 \\ \hline
        BioAgents & 0.121 & 0.012 & 0.071 & 0.086
        \\ \hline

    \end{tabular}
    \caption{Benchmarking on 71 Biostars Question-Answer (QA) Pairs}
    \label{benchmarking}
\end{table}

Outputs from BioAgents and our Human Experts for the easy and medium workflows are displayed below:

\begin{figure}[ht]
  \centering
  \begin{minipage}[b]{0.44\textwidth}
    \centering
\fbox{\includegraphics[width=\textwidth]{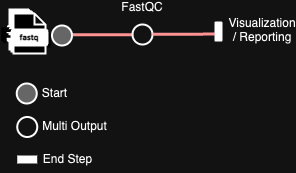}}
    \caption{Multi-Agent System on the Easy Workflow}
    \label{easy_system}
  \end{minipage}
  \hspace{4pt}
  \begin{minipage}[b]{0.44\textwidth}
    \centering
    \fbox{\includegraphics[width=\textwidth]{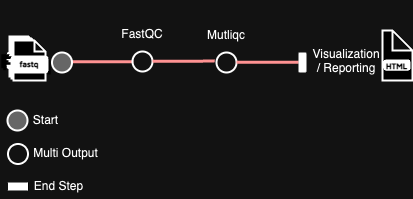}}
    \caption{Experts on Easy Workflow}
    \label{easy_expert}
  \end{minipage}
\end{figure}

\begin{figure}[ht]
  \centering
  \begin{minipage}[b]{0.44\textwidth}
    \centering
\fbox{\includegraphics[width=\textwidth]{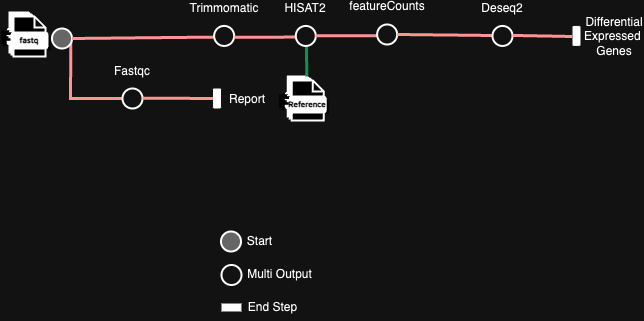}}
    \caption{Multi-Agent System on the Medium Workflow}
    \label{med_system}
  \end{minipage}
  \hspace{4pt}
  \begin{minipage}[b]{0.44\textwidth}
    \centering
    \fbox{\includegraphics[width=\textwidth]{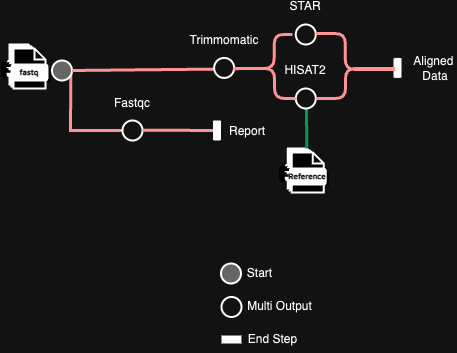}}
    \caption{Experts on Medium Workflow}
    \label{med_expert}
  \end{minipage}
\end{figure}

\bibliographystyle{acm}
\bibliography{references}

\end{document}